\documentclass[11pt]{article}

\usepackage[final]{acl}
\usepackage{times}
\usepackage{latexsym}
\usepackage{multirow}
\usepackage[T1]{fontenc}

\usepackage[utf8]{inputenc}

\usepackage{microtype}

\usepackage{inconsolata}

\usepackage{graphicx}
\usepackage{algorithm}
\usepackage{algorithmic}
\usepackage{xcolor}
\usepackage{enumitem}
\usepackage{amsmath}
\usepackage{tcolorbox}
\usepackage{amsfonts}
\usepackage{nameref}
\usepackage{xcolor}
\usepackage{colortbl}

\usepackage{newfloat}
\usepackage{listings}
\usepackage{tabularx}
\usepackage{booktabs}  
\usepackage{caption}   
\title{Mobile-R1: Towards Interactive Capability for VLM-Based Mobile Agent via Systematic Training}

\author{First Author \\
  Affiliation / Address line 1 \\
  Affiliation / Address line 2 \\
  Affiliation / Address line 3 \\
  \texttt{email@domain} \\\And
  Second Author \\
  Affiliation / Address line 1 \\
  Affiliation / Address line 2 \\
  Affiliation / Address line 3 \\
  \texttt{email@domain} \\}
\author{
  \textbf{Jihao Gu}\thanks{Equal contribution, Random order.}\thanks{Work done during an internship at Alibaba Group.}\textsuperscript{1}, 
  \textbf{Qihang Ai}\footnotemark[1]\footnotemark[2]\textsuperscript{1}, 
  \textbf{Yingyao Wang}\footnotemark[1]\textsuperscript{1}, 
  \textbf{Pi Bu}\footnotemark[1]\textsuperscript{1}, 
  \textbf{Jingxuan Xing}\footnotemark[1]\textsuperscript{3}, \\
  \textbf{Yue Cao}\textsuperscript{1},
  \textbf{Zekun Zhu}\textsuperscript{3},
  \textbf{Wei Jiang}\textsuperscript{3},
  \textbf{Ziming Wang}\textsuperscript{1},\\
  \textbf{Yingxiu Zhao}\textsuperscript{1},
  \textbf{Ming-Liang Zhang}\textsuperscript{1},
  \textbf{Jun Song}\thanks{Corresponding Author.}\textsuperscript{1,2},
  \textbf{Yuning Jiang}\footnotemark[3]\textsuperscript{3}, 
  \textbf{Bo Zheng}\textsuperscript{1,2,3} \\
  \textsuperscript{1}Future Living Lab, Alibaba Group
  \textsuperscript{2}Alibaba Group Holding Limited \\
  \textsuperscript{3}Taobao \& Tmall Group of Alibaba \\
  {\{gujihao.gjh,  aiqihang.aqh, wangyingyao.wyy, bupi.wj, jsong.sj\}}@taobao.com
}

\begin{document}
\maketitle
\begin{abstract}
Vision-language model-based mobile agents have gained the ability to understand complex instructions and mobile screenshots
, benefiting from reinforcement learning paradigms like Group Relative Policy Optimization (GRPO). 
However, existing approaches centers on offline training or local action-level rewards often trap agents in local optima, hindering effective exploration and error correction with the environment. 
Crucially, we find that directly applying task-level rewards often leads to convergence difficulties due to the sparse nature of GUI interactions. 
To address these challenges, we present \textbf{Mobile-R1}, a systematic training recipe that bridges atomic action execution and strategic task completion. 
We propose a hierarchical curriculum consisting of three stages: (1) format alignment for reasoning structure, (2) on-policy exploration with verifiable action feedback to ground basic execution, and (3) multi-turn task-level training with realistic environment to unlock exploration and self-correction. 
This hierarchical strategy effectively bootstraps the agent, significantly enhancing its capability for exploration and self-correction (the ``Eureka'' moments). Furthermore, addressing the critical scarcity of diverse GUI data in non-English ecosystems, we contribute a comprehensive Chinese mobile dataset covering 28 applications with 24,521 high-quality manual annotations, and establish a rigorous benchmark with 500 trajectories. We will open source all resources, including the dataset, benchmark, model weight, and codes: \url{https://mobile-r1.github.io/Mobile-R1/}.
\end{abstract}

\section{Introduction}

Vision Language Model (VLM)-based agents have the capability to effectively integrate textual instructions with visual inputs, allowing them to devise comprehensive operational strategies and execute actions for complex tasks~\cite{li2025summary,gu2025simplevqa}. These agents not only comprehend intricate instructions but also engage in multi-turn planning~\cite{nguyen2024gui,huang2024understanding} and interact with external tools or environments~\cite{yuan2024easytool,shao2023character}, making them particularly well-suited for autonomous operation on mobile devices. Specifically, VLM-based mobile agents are driven by textual instructions, understand screenshots of mobile screens, and generate multi-turn actions to accomplish the task goals required by the instructions.

Several pioneers have explored relevant technologies. For instance, the AppAgent~\cite{li2024appagent} and Mobile-Agent series~\cite{mobileagent} introduced multi-modal agents, while UI-TARS~\cite{qin2025ui} excels in comprehending screenshot content and navigating graphical user interfaces (GUIs). With the advancement of foundational models, such as Qwen2.5-VL~\cite{bai2025qwen25vl}, mobile agents are demonstrating even greater potential. 
Recently, studies like UI-R1~\cite{ui-r1} and GUI-R1~\cite{luo2025guir1} draw inspiration from DeepSeek-R1~\cite{guo2025deepseek}, attempting to use Group Relative Policy Optimization (GRPO) to guide the model's thinking and reasoning about the environment and actions.
These above methods can adapt to immediate observations but struggle with the changing mobile environments, due to their reliance on action-level rewards that only guide the agent to predict the best single action at each step.

\begin{figure*}[t]
    \centering
  \includegraphics[width=1\linewidth]{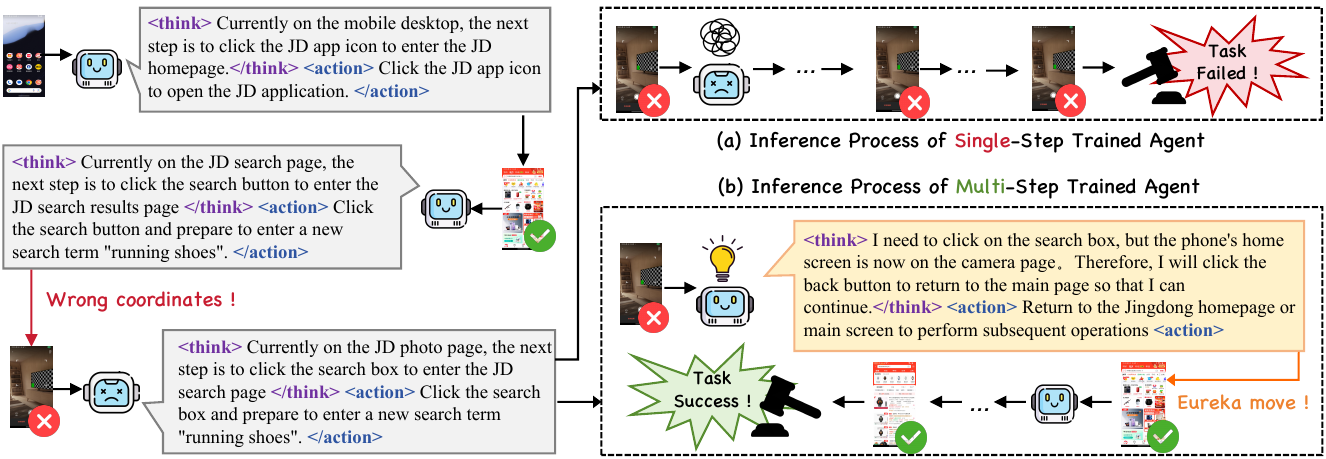}
    \caption{Comparison of Action-Level and Task-Level Rewards. Figure (a) illustrates an action-induced agent that is encouraged to ``think longer'' before selecting a single-step action. In contrast, Figure (b) depicts an agent trained at the task-level rewards, which explores and adjusts its trajectory over multi-turn interactions with the environment.}
    \label{fig:intro}
    \vspace{-0.5cm}
\end{figure*}

Notably, we emphasize that a mobile agent should engage in dynamic, multi-turn interactions with the mobile environment. Therefore, employing action-guided SFT or RL training alone is suboptimal and merely a temporary solution. As illustrated in Figure \ref{fig:intro}(a), the action-induced agent is encouraged to ``think longer'' merely to select a precise single-step action. 
This excessive supervision of action-level rewards can easily lead to local optima, significantly diminishing the model's capacity for exploration and self-correction. Conversely, multi-turn task-oriented learning is the ideal approach, as shown in Figure \ref{fig:intro}(b). 
However, directly applying task-level rewards often leads to training instability due to the sparse nature of success signals in complex GUIs. Successful training thus depends on satisfying three core requirements: \textbf{1) Multi-turn Trajectories:} the agent must incorporate historical trajectories to optimize for task completion rather than single turns. \textbf{2) Verifiable On-policy Exploration:} before attempting long-horizon tasks, the agent must first stabilize its interaction capability through online exploration with verifiable feedback, avoiding the cold-start failure of sparse rewards. \textbf{3) Trajectory Correction:} the agent should possess long-term planning and error reflection abilities to prevent getting trapped in local dilemmas.

In this paper, we strive to develop a systematic training recipe including realitic environment reinforcement learning, namely \textbf{Mobile-R1}, for the VLM-based mobile agent to acquire the interactive capability. 
To ensure stable training and bridge the gap between action execution and task planning, we propose a robust three-stage hierarchical curriculum: format finetuning, on-policy exploration with verifiable feedback, and task-level training. 
In Stage 1, we gather high-quality action trajectory samples, including self-correction trajectories, to enable the model to learn error correction formats (Cold Start). 
Thereafter, Stage 2 employs on-policy GRPO with **verifiable action feedback** (i.e., combining ground-truth matching with format constraints) to ground the agent's execution capability.
In Stage 3, multi-turn GRPO training is conducted, focusing on task-level rewards that are applied to the entire trajectory. This stage is designed to encourage dynamic exploration within multi-turn interactions. 
Moreover, we introduce a new benchmark encompassing 500 trajectories with 1842 steps in total as well as a high-Quality trajectory dataset, addressing the under-representation of the Chinese ecosystem in existing mobile agent research. Our method demonstrates superior performance on this benchmark. Surprisingly, we discover that our agent is capable of correcting itself from an incorrect state back to the correct action (called the \textit{eureka move}), demonstrating the emergence of reasoning capabilities through our multi-turn online reinforcement learning.

The main contributions are as follows: 
\begin{itemize}[leftmargin=*]
\item{\textbf{Comprehensive Chinese Benchmark.} We introduce a challenging mobile agent benchmark that includes 500 trajectories with human annotation, filling the gap in non-English GUI evaluation.}
\item{\textbf{High-Quality Trajectory Dataset.} We contribute a high-quality dataset featuring 4,635 manually annotated trajectories with 24,521 steps in total, which facilitates robust VLM-based agent training.}
\item{\textbf{Robust Training Recipe for Mobile-R1.} We present a systematic three-stage training strategy that effectively bridges atomic action execution and task-level planning. Enabling multi-turns of interaction between the mobile agent and the environment. Experiments confirm this strategy enables critical self-correction behaviors.}
\item{\textbf{Open Source Resources.} We will open source all our resources, including the dataset, benchmark, model weight, and codes\footnote{\url{https://mobile-r1.github.io/Mobile-R1/}}.}
\end{itemize}
\begin{table*}[t]
\centering
\renewcommand{\arraystretch}{1.05}
\vspace{-0.3cm}
\resizebox{0.85\textwidth}{!}{
\begin{tabular}{l|c|c|c|c}
\toprule
\textbf{Agents} & \textbf{Backbone} & \textbf{Training Strategy} & \textbf{Reward Source} & \textbf{Environment} \\ \hline
Agent Q~\citep{agentq}&  LLM-based  &    DPO     &     Action-level        &         Off-line         \\ \hline
 ARPO~\citep{lu2025arpo}&  VLM-based&    GRPO&     Task-level&         On-line\\ \hline
UI-TARS \citep{qin2025ui} &  VLM-based  &    DPO     &     Action-level       &         Off-line         \\ \hline
GUI-R1 \citep{luo2025guir1} &  VLM-based  &    GRPO&     Action-level        &         On-line         \\ \hline
AgentCPM \citep{zhang2025agentcpm} &  VLM-based  &    GRPO&     Action-level        &         On-line         \\ \hline
GUI-Critic-R1 \citep{gui-critic-r1} &  VLM-based  &    GRPO&     Action-level        &         On-line         \\ \hline
InfiGUI-R1 \citep{liu2025infigui} &  VLM-based  &    GRPO&     Action-level        &         On-line         \\ \hline
 UI-R1~\citep{ui-r1}& VLM-based& GRPO& Action-level &  On-line\\ \hline
\textbf{Mobile-R1 (Ours)}  &   VLM-based & GRPO  &  Action- \& Task-level &  On-line \\
\bottomrule
\end{tabular}} 
\caption{Comparison of several existing RL-based agent. Among them, ``Reward Source'' refers the agent's reward for executing an single action (i.e., action-level), or completing a task (i.e., task-level).
}
\vspace{-0.5cm}
\label{table:intro}

\end{table*}
\section{Related Work}
\subsection{Mobile Agent Framework}
Graphical User Interface (GUI) agents are designed to operate in digital environments with graphical elements such as buttons and images. Their applications span web navigation, mobile app interactions, and desktop automation~\cite{chen2025combatvla}. 
In the mobile agent, work has evolved from API-based frameworks using commercial models to open-source, end-to-end frameworks. Earlier API-based frameworks such as the AppAgent series~\cite{zhang2025appagent,li2024appagent} and the Mobile-Agent series~\cite{mobileagent,mobileagent_v2,mobileagente} used commercial models like GPT for planning and prediction, relying on complex workflows; MobileSteward~\cite{liu2025mobilesteward} extends this direction to cross-app instructions via self-evolving app-oriented agents. Recent advancements in open-source VLM have led to training these models on GUI-specific data. For instance, UI-TARS~\cite{qin2025ui} continuously trains Qwen-2-VL~\cite{wang2024qwen2vl} models, specifically the 7B and 72B variants, on approximately 50 billion tokens. 
ShowUI~\cite{lin2025showui} enhances Qwen2-VL-2B using UI-guided token selection and high-quality GUI datasets. UI-R1~\cite{ui-r1} explores rule-based reinforcement learning to boost VLMs' reasoning, while InquireMobile~\cite{ai2025inquiremobile}
further teaches the agent to actively request human assistance under
the same RL paradigm.

Progress along these lines is increasingly assessed by dedicated benchmarks
such as Mobile-bench~\cite{deng2024mobile} and
AndroidLens~\cite{cao2025androidlens}, which probe LLM-driven workflows
and long-latency tasks with nested sub-targets, respectively.

\subsection{Visual Reasoning Model}
DeepSeek R1~\cite{guo2025deepseek} showed that reinforcement learning with rule-based incentives helps large language models (LLMs) develop unique reasoning skills. Researchers are expanding this to multi-modal reasoning. VisualThinker-R1-Zero~\cite{zhou2025r1} is the first to achieve enhanced visual reasoning with a non-SFT 2B model. Visual Reinforcement Fine-Tuning (Visual-RFT)~\cite{liu2025visual} targets visual tasks, including image classification, object detection, and reasoning grounding. Skywork R1~\cite{peng2025skywork} uses multi-modal transfer to improve R1-series LLMs for visual tasks, combining SFT with reinforcement learning (i.e., GRPO) for better cross-modal reasoning. Notable works include R1-OneVision~\cite{yang2025r1-ov} and R1-V~\cite{chen2025r1v}, with R1-V exploring Reinforcement Learning with Verifiable Reward (RLVR) to boost VLMs' visual reasoning. It demonstrated that RLVR methods exhibit strong out-of-distribution generalization, whereas SFT excels in training domain tasks~\cite{chen2025r1vblog}.
\section{Trajectory Dataset} \label{sec:dataset}
\begin{figure}
    \centering
    \includegraphics[width=1\linewidth]{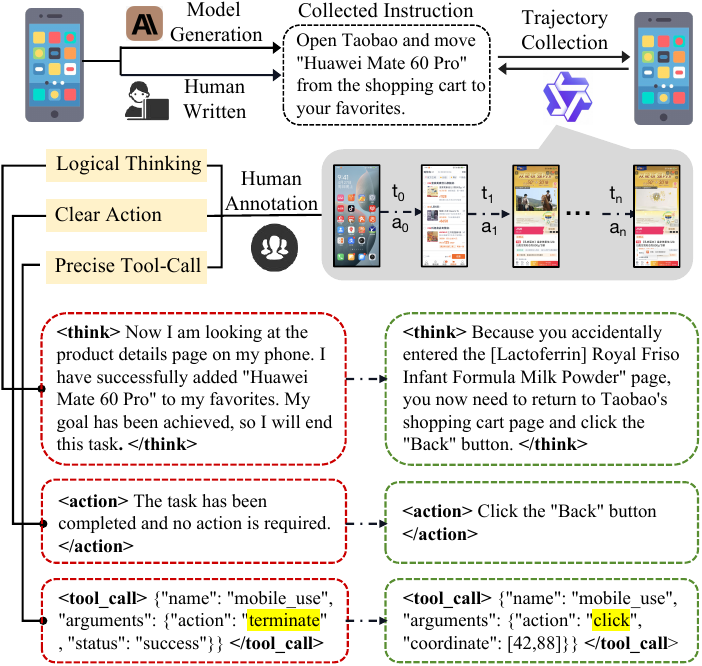}
    \caption{Pipeline of data collection.}
    \label{fig:datacollection}
    \vspace{-0.5cm}
\end{figure}

To address the scarcity of data with explicit reasoning supervision and the under-representation of non-English ecosystems, we first constructed a high-quality dataset comprised of 4,635 trajectories with 24,521 manually annotated steps, supporting the community's efforts in training and evaluating powerful agents. Unlike prior datasets that only focus on action outcomes, ours explicitly captures the intermediate reasoning process within complex Chinese mobile applications.
The pipeline of data collection is shown in Figure \ref{fig:datacollection}, which is divided into trajectory collection and trajectory annotation.

\subsection{Trajectory Collection}
We first selected 28 Chinese mobile apps, including both commercial and system types\footnote{All apps and prompts are provided in the Appendix.}. We created a diverse set of instructions for each app through manual crafting of common tasks and automatic generation by Claude 3.5 Sonnet~\cite{anthropic2024claude}. These instructions were manually reviewed, and any unreasonable ones were removed, leaving 1,510 instructions. Thereafter, we used the Qwen2.5-VL-3B~\cite{bai2025qwen25vl} model as the mobile agent to execute these tasks, with a maximum of 25 steps, allowing multiple executions per instruction to simulate different mobile states for the same task. In this process, we gather a total of 4,635 raw action execution trajectories.

\begin{figure*}
    \centering
  \includegraphics[width=1\linewidth]{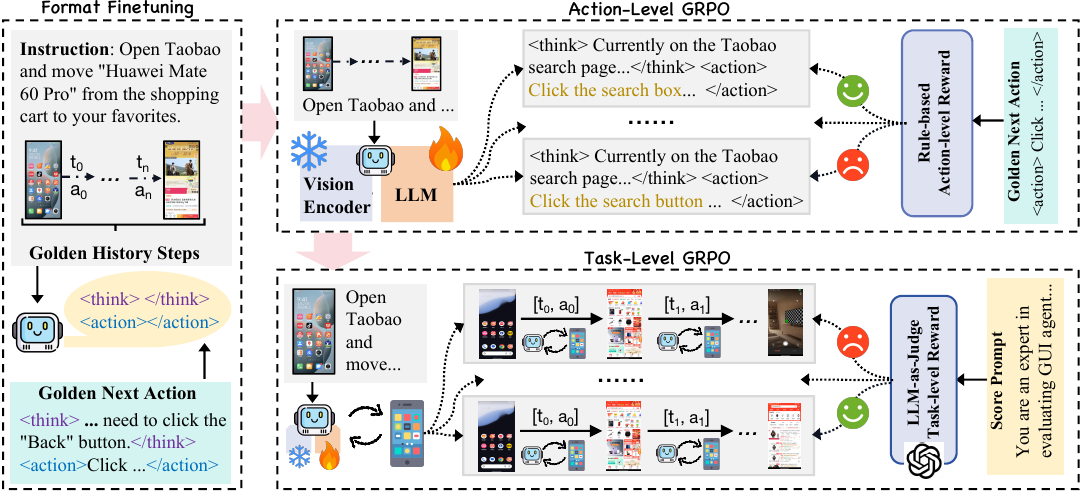}
    \caption{Our training framework consists of three stages: initial format finetuning, online training via action-level reward, followed by online training via task-level reward based on multi-turn trajectories.}
    \label{fig:framework}
    \vspace{-0.3cm}
\end{figure*}

\subsection{Trajectory Annotation}
\label{subsec:annotation}
As shown in Figure \ref{fig:datacollection}, the action trajectory annotation process consists of a three step: logical thinking, clear action, and precise tool-calling. 

\paragraph{Logical Thinking}
In this step, annotations ensure logical coherence and decision-making throughout the action trajectory. Annotators evaluate the model's ``thinking'' for errors or redundancies using a format: 
\\
\vspace{-0.3cm}
\begin{center}
\fcolorbox{black}{gray!10}{\parbox{0.95\linewidth}{
\texttt{<think>}{\color{red}Currently on the phone home screen}, {\color{blue}the next step is to click the Taobao app} {\color{purple} to enter Taobao}.\texttt{<think>}
}}
\end{center}
The {\color{red}red} part indicates the current state or interface,  the {\color{blue}blue} part indicates the next action and the {\color{purple}purple} part indicates the goal of the action.
Correct but redundant or incorrect data is rewritten.

\paragraph{Clear Action}
This stage involves clarifying instructions to ensure they are clear and explicit. These instructions guide actions according to ANDROIDCONTROL~\cite{androidcontrol}.

\paragraph{Precise Tool-Calling}
Through careful annotation in the previous two stages, we obtained thinking and basic instructional tasks. Here, "tool-calling" refers to mapping the action’s natural language description into our standardized action space. Thereafter, annotators evaluate if actions are effective using the current screenshot and action descriptions. Correct tool-calls are kept, while incorrect ones are corrected.

\subsection{Dataset Statistics}
\begin{figure}
    \centering
    \includegraphics[width=1\linewidth]{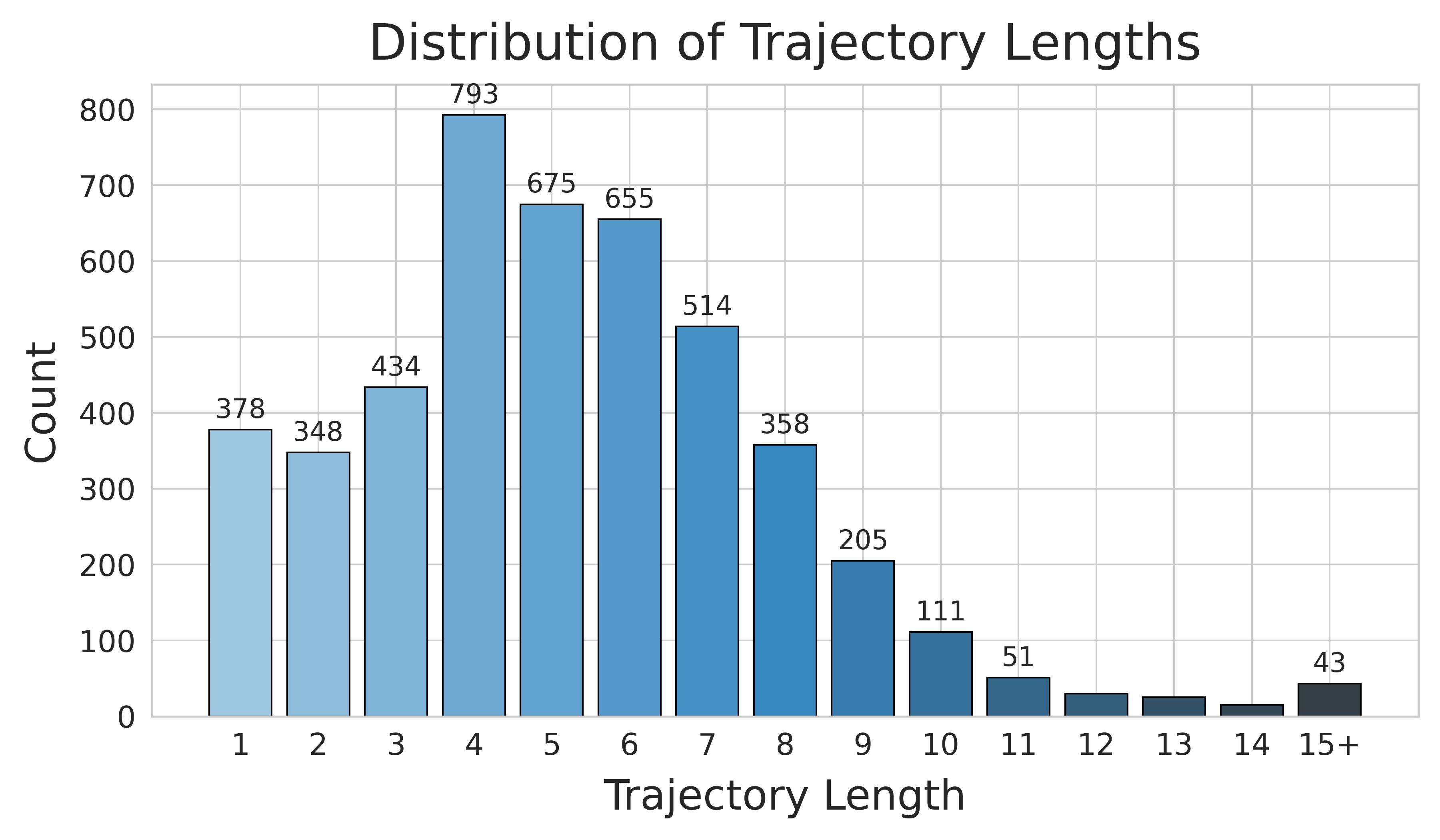}
    \vspace{-0.5cm}
    \caption{Distribution of trajectory length.}
    \label{fig:trajectory_stat}
    \vspace{-0.5cm}
\end{figure}

The overview of the dataset is shown in Table~\ref{tab:data_stat}. The dataset consists of 4,635 high-quality, fine-grained mobile interaction trajectories, characterised by rigorous human verification and diverse task complexity. The distribution of the trajectory length is detailed in Figure~\ref{fig:trajectory_stat}.
\begin{table}[h]
    \centering
    \resizebox{0.48\textwidth}{!}{
    \begin{tabular}{lcccc}
        \toprule
        \textbf{Type} & App& Instruction& Trajectory& Data\\
        \midrule
        \textbf{Number}& 28& 1,510& 4,635& 24,521\\
        \bottomrule
    \end{tabular}
    }
    \caption{Overview of our trajectory dataset.}
    \label{tab:data_stat}
    \vspace{-0.5cm}
\end{table}

\section{Our Mobile-R1}
As shown in Figure \ref{fig:framework}, Mobile-R1 employs a curriculum learning framework designed to bootstrap reasoning in sparse-reward environments. The training recipe consists of three progressive stages: 1) Format Alignment (Warm-up) using the CoT data described in 
Section~\ref{sec:dataset}, 
2) On-Policy Exploration with Verifiable Feedback to ground atomic execution, and 3) Multi-Turn Task-Level Optimization to unlock long-horizon planning and self-correction.
\subsection{Preliminary}
\subsubsection{Task Definition}
Mobile agents are driven by textual instructions, understand screenshots of mobile screens, and multi-turn generate action to accomplish the task goals. Instructions are divided into task level (``create an event for tomorrow at 2 pm'') and action level (``click the icon in the top left corner'').

\subsubsection{Response Format}
Following the pioneers, we format response as \texttt{<think>}, \texttt{<action>}, \texttt{<tool\_call>}.

\subsubsection{Action Space}
We adopt a unified action space to ensure that all task-level instructions can be decomposed into a sequence of atomic actions via \texttt{<tool\_call>}. There are eight actions: key, click, swipe, long press, type, system button, terminate, and wait\footnote{The definitions can be found in the Appendix.}.

\subsubsection{Group Relative Policy Optimization}
GRPO \cite{shao2024deepseekmath} builds on the Proximal Policy Optimization (PPO) by using group-normalized token-level advantages. This method tackles the issue of sparse or high-variance rewards in LLMs without needing a value function or critic.

Given a batch of $G$ generated responses $\{o_i\}_{i=1}^G$ from a query, where each response $o_i = (o_i(1), \dots, o_i(|o_i|))$, the GRPO objective function is defined as:
\begin{equation}
\label{eq:grpo_objective}
\footnotesize 
\begin{split} 
&J_{\text{GRPO}}(\theta) = \frac{1}{G} \sum_{i=1}^{G} \frac{1}{|o_i|} \sum_{t=1}^{|o_i|} \min \left[ \frac{\pi_{\theta}(o_i(t)|o_i,<t)}{\pi_{\text{old}}(o_i(t)|o_i,<t)}\right. \\ 
&\quad \left.  \hat{A}_{i,t},\text{clip}\left(\frac{\pi_{\theta}(o_i(t)|o_i,<t)}{\pi_{\text{old}}(o_i(t)|o_i,<t)}, 1-\epsilon, 1+\epsilon\right) \hat{A}_{i,t} \right]
\end{split} 
\end{equation}
where: $\pi_{\theta}(o_i(t)|o_i,<t)$ and $\pi_{\text{old}}(o_i(t)|o_i,<t)$ are the probabilities of generating token $o_i(t)$ under the current and old policies, respectively. $\epsilon$ is the clipping hyperparameter for the probability ratio. $\hat{A}_{i,t}$ is the group-normalized advantage for token $o_i(t)$ in response $o_i$:
\begin{equation}
\label{eq:group_norm_advantage}
\hat{A}_{i,t} = \frac{r_i - \mu}{\sigma},
\end{equation}
with $r_i$ being the total reward for response $o_i$, and $\mu, \sigma$ being the mean and standard deviation of all rewards $\{r_j\}_{j=1}^G$ within the current batch.

\subsection{Stage1: Format Finetuning}
This stage aims to train the model to predict the next action from instructions and operation history. To equip the model with this fundamental capability, we start with initial Supervised Fine-Tuning (SFT) as cold start using the action-level instructions from
Section~\ref{sec:dataset}. 
This essential stage builds a strong link between user intent, GUI state, and actions, supporting later reinforcement learning.

\subsection{Stage2: On-Policy Exploration with Verifiable Feedback}
Subsequently, to prevent the "cold-start" failure common in sparse-reward settings, the model was trained using GRPO through action-level rewards with verifiable feedback (derived from ground-truth annotations) as a dense reward signal. The reward function at this stage is a combination of two components: a verifiable action reward and a format reward, as follows:
\begin{equation}
\label{eq:total_reward}
R_{action} = R_{Act} + R_{Fmt},
\end{equation}
where $R_{Act}$ quantifies the correctness of the executed action, and $R_{Fmt}$ ensures the format integrity of the generated output.

\subsubsection*{Action-Level Reward ($R_{Act}$)}
$R_{Act}$ assesses the correctness of the action prediction, with its calculation varying based on the action type.
\begin{itemize}
    \item For coordinate-based actions (e.g., click, swipe), $R_{Act}$ is 1 if the predicted coordinate $C=[x,y]$ falls within the ground truth bounding box $B=[x_1,y_1,x_2,y_2]$ of the target GUI element; otherwise, it is 0. This ensures precise spatial interaction.
    \item For non-coordinate actions (e.g., `type(text=)'), $R_{Act}$ is 1 if the predicted action or its argument (e.g., the text string to type, the specific `back' command) exactly matches the ground truth; otherwise, it is 0. This guarantees faithful command execution.
\end{itemize}

\subsubsection*{Format Reward ($R_{Fmt}$)}
$R_{Fmt}$ incentivizes the model to produce structured, interpretable outputs. 
\begin{itemize}
    \item \texttt{<think>}: The internal reasoning processes.
    \item \texttt{<action>}: The immediate next action to be executed.
    \item \texttt{<tool\_call>}: The final answer or tool/API invocation for the current step.
\end{itemize}
$R_{Fmt}$ is a binary reward (1 for full compliance, 0 otherwise) that enforces adherence to these tagging and structural requirements, ensuring well-formed and semantically organized responses.

\subsection{Stage3: Task-level Online Training}
\label{sec:stage3}
While Stage 2 ensures atomic precision, it encourages greedy local optima. In Stage 3, we transition to multi-step task-level GRPO to enable strategic exploration. Here, the agent receives a reward upon task completion, forcing it to credit assignment over long horizons, enabling the robuteness of agent and emergence of "Eureka" moments.

Firstly, we define this multi-turn interaction problem as a Markov decision process~\cite{lu2025arpo}.

Each trajectory $\tau$ is a sequence of observations $s_t$ (representing mobile screenshot), agent actions $a_t$ (including thinking, action and tool\_use), and a scalar reward $R$ obtained upon trajectory completion. The core objective is to train a policy $\pi_\theta$ that maximizes the accumulated rewards over these interaction sequences, where $a_t$ is sampled from the policy conditioned on the preceding state-action history:
\begin{equation}
\label{eq:trajectory_def}
\begin{split}
&\tau = \{s_t, a_t\}_{t=0}^{T-1}, \quad \text{where} \\
&a_t \sim \pi_\theta (\{s_i\}_{i=\max(0, t-W)}^{t-1}, \{a_i\}_{i=0}^{t-1}).
\end{split}
\end{equation}

We define $W$ as the sliding window size, which controls the maximum number of observations ($s_i$) considered.
It only caps the number of \emph{historical screenshots} $\{s_i\}$ fed to the model, which is a VRAM constraint. The full textual history ($\langle$think$\rangle$, $\langle$action$\rangle$, $\langle$tool\_call$\rangle$) is always preserved, so the agent can still reason over steps outside the visual window (e.g., recognizing an earlier mis-click). The specific value of $W$ is given in Appendix~\ref{appendix: training settings}.

The reward $R_{task}$ of this stage is composed of two components: a Format Reward ($R_{Fmt}$) and a Trajectory-Level Reward ($R_{Traj}$), formulated as:
\begin{equation}
\label{eq:multi_turn_reward}
R_{task} = R_{Fmt} + R_{Traj}.
\end{equation}

\subsubsection*{Format Reward ($R_{Fmt}$)}
Differing from Stage 2, in this stage, $R_{Fmt}$ for the entire trajectory is obtained by averaging the format reward of all actions. Moreover, $R_{Fmt}$ is set to $[-1, 1]$ to impose stricter penalties for errors.

\subsubsection*{Trajectory-Level Reward ($R_{Traj}$)}
To obtain a comprehensive evaluation signal for multi-turn interactions, an external, high-fidelity MLLM, GPT-4o \cite{openai2023gpt}, is employed to assign a scalar reward score to the entire historical interaction trajectory $\tau = (s_0, a_0, \dots, a_n)$. 

Drawing inspiration from prior work \cite{sun2024genesis}, we establish two primary scoring criteria for GPT-4o\footnote{The version we used is GPT-4o-0806. Prompts provided to GPT-4o can be found in the Appendix.}:
\begin{itemize}
    \item Trajectory Coherence: This checks if steps and actions consistently follow the target instruction, actions are clear and specific, and if there are no unnecessary steps.
    \item Task Completion: This evaluates if the task is fully completed, all necessary interactions are made, and errors are handled properly.
\end{itemize}
The 5-level scoring rubric is applied by GPT-4o, yielding a final score within the range $[0, 1]$. 
\section{Experiment}

\subsection{Implementation Details}
\subsubsection{Virtual Environment}
The Android Studio emulator serves as our primary mobile GUI interactive environment. A local monitoring script runs alongside the emulator, actively managing the interaction loop.

\subsubsection{Datasets and Benchmark}
In the first two stages, we utilized 1,000 and 3,459 trajectory samples. In the third stage, we trained using only five frequently used Android apps—Jingdong, Pinduoduo, Taobao, Fliggy, and Bilibili—creating 20 unique task trajectories per app, totaling 100. 
For our evaluation benchmark, we separate 500 human-annotated trajectories with 1,842 steps from the dataset, of which 225 trajectories are specifically from long-tail unseen applications to better evaluate generalization. We also evaluated English benchmark AndroidControl\cite{li2024effects} and GUI-Odyssey\cite{lu2024gui}.


\subsubsection{Training Settings}
Our experiments utilize Qwen2.5-VL-3B as the base model, with GRPO implementation (including for trajectory-level interaction training) adapted from the open-r1 framework \cite{openr1}. For hyperparameter settings, Stage 1: train for 2 epochs at a learning rate of $1 \times 10^{-4}$. Stage 2: train for 2 epochs at $1 \times 10^{-7}$ with 8 rollouts and a temperature of 1 for exploration. Stage 3: train for 2 epochs at $1 \times 10^{-6}$ with a temperature of 1.

More detailed hyperparameter configurations are listed in the Appendix~\ref{appendix: training settings}.

\subsubsection{Baselines}
Qwen2.5-VL-3B \cite{bai2025qwen25vl}, UI-R1-3B, UI-R1-3B-E \cite{ui-r1}, GUI-R1-3B, GUI-R1-7B \cite{luo2025guir1}, AgentCPM-8B \cite{zhang2025agentcpm} are baselines.

\begin{table}[t]
    \centering
    \resizebox{0.48\textwidth}{!}{
        \small
        \setlength{\tabcolsep}{3.8pt} 
        \begin{tabular}{l r r r r r}
\toprule
Model                       & Acc. & Task Succ. & Tail Succ. & Avg Err.$\downarrow$ \\
\midrule
Qwen2.5-VL-3B & 54.49 & 7.20 & 16.33  & 651 \\
Qwen2.5-VL-7B & 63.46 &	12.80 &	21.60 &	523 \\
Qwen2.5-VL-32B & 75.90& 30.40& -& 	280\\
UI-R1-3B&	56.13&	17.20&	-&	451\\
UI-R1-3B-E&	59.12&	9.40&	-&	473\\
GUI-R1-3B&	61.29&	12.00&	-&	461\\
GUI-R1-7B&	71.72&	\textbf{32.60}&	-&	298\\
AgentCPM-8B& 71.65&	30.00&	-&	338 \\
\midrule
\rowcolor[gray]{0.9}\textbf{Mobile-R1 (Ours)} & \textbf{78.55} & \underline{30.60} & \textbf{37.40} & \textbf{241} \\
\rowcolor[gray]{0.9}Stage1 \& Stage2 & \underline{77.69} & 29.40 & \textbf{36.00} & \underline{255} \\
\rowcolor[gray]{0.9}Stage1 & 75.68 &	24.40 &	29.80 &	280 \\
\bottomrule
\end{tabular}}
        \captionof{table}{Overall Performance Comparison. \textbf{Bold} and \underline{underline} indicate the best and second-best results.}
\label{tab:performance_all_centered}
\vspace{-0.5cm}
\end{table}


\subsubsection{Evaluation Metrics}
We evaluate the model's performance using the following metrics:
\begin{itemize}
    \item \textbf{Accuracy (Acc.):} The probability of correctly performing each step in a trajectory, correct if both format and action match definitions $R_F$ and $R_{Act}$.
    \item \textbf{Task Success Ratio (Task Succ.):} The probability of a complete trajectory being executed entirely correctly.
    \item \textbf{Tail Success Ratio (Tail Succ):} The probability that the task within a trajectory is ultimately completed successfully, regardless of intermediate errors or deviations.
    \item \textbf{Action Argument Error Number (Avg Err.):} The count of errors of incorrect action.
\end{itemize}


\begin{table*}[t]
\resizebox{\textwidth}{!}{
\small
\centering
\begin{tabular}{lcccccccc}
\toprule
\cmidrule(lr){1-9}
\multirow{2}{*}{Model}
& \multicolumn{2}{c}{Android Low} & \multicolumn{2}{c}{Android High} & \multicolumn{2}{c}{Odyssey} & \multicolumn{2}{c}{AITZ}\\
\cmidrule(lr){2-3} \cmidrule(lr){4-5} \cmidrule(lr){6-7} \cmidrule(lr){8-9}
& TM & EM & TM & EM & TM & EM & TM & EM\\
\midrule
Qwen2.5-VL-7B & \textbf{94.1} & 85.0   & 75.1 & 62.9 & 59.54 & 46.28 & --             & --             \\
OS-Genesis-7B & 90.7          & 74.2   & 65.9 & 44.4 & 11.67 & 3.63  & 19.98          & 8.45           \\
OS-Atlas-7B   & 73.0          & 67.3   & 70.4 & 56.5 & \textbf{91.83} & \textbf{76.76} & 74.13 & 58.45 \\
Aguvis-7B     & 93.9          & 89.4   & 65.6 & 54.2 & 26.71 & 13.54 & 35.71          & 18.99          \\
Odyssey-7B    & 65.1          & 39.2   & 58.8 & 32.7 & 90.83 & 73.67 & 59.17          & 31.60          \\
\rowcolor[gray]{0.9}\textbf{Mobile-R1(3B)} & 93.5 & \textbf{87.1} & \textbf{76.5} & \textbf{65.2} & 75.24 & 53.11 & \textbf{77.05} & \textbf{60.50}\\
\bottomrule
\end{tabular}}
\caption{Performance Comparison on Android Control (Android), GUI-Odyssey (Odyssey), and AITZ~\cite{zhang2024aitz}. \textbf{Bold} indicate the best results. ``--'' denotes that AITZ numbers are not reported for the base model.}
\label{tab:performance_android}
\vspace{-0.4cm}
\end{table*}

\subsection{Experimental Result}

We evaluated all models on our benchmark, with experimental results shown in Table~\ref{tab:performance_all_centered}. 
We have the following observations: 
1) Our Mobile-R1 outperformed all baselines, achieving an accuracy of 78.55, 2.7 points higher than the best baseline. With Stage 3 training, the Mobile-R1's Task Success Ratio increased by 1.2 points over the Stage 1 \& 2 model, benefiting from the task-level GRPO. 
2) Our Stage 1 \& 2 allows the Qwen2.5-VL-3B model to surpass its standard version and outperform baselines, highlighting the importance of action- and task-level rewards.

Notably, to further verify the necessity of our hierarchical training recipe, we conducted an experiment that directly proceeds from Stage 1 to Stage 3. Skipping Stage 2 causes the reward to plateau throughout Stage 3. Without the dense supervision provided by verifiable action rewards in Stage 2, the agent fails to accumulate meaningful task-level signals under the sparse-reward setting and thus cannot improve upon the SFT baseline. We have tested reward-versus-steps curve of our Stage 3 shown in Appendix~\ref{appendix: task-level training}.

Moreover, we extended our evaluation to the English benchmark, Android Control and GUI-Odyssey with related works \cite{sun2024genesis, wu2024atlas, xu2024aguvis, lu2024gui}. These benchmarks utilize two standard metrics: Type Match (TM), which verifies if the predicted action type matches the ground truth, and Exact Match (EM), which additionally requires all parameters to be correct. Table \ref{tab:performance_android} showed that, for Android Control, our 3B model demonstrated significantly superior performance compared to a range of 7B models. The sole exception was a slightly lower score on the TM-Low metric, which we attribute to the inherent limitations of a 3B model in advanced instruction-following. 
For GUI-Odyssey, our model achieved robust performance with only 3B parameters.
All results are second only to the Odyssey-7B and Aguvis-7B models that underwent targeted training.
Overall, these results underscore our model's robust and competitive performance on English-language tasks.

Moreover, we extended our evaluation to the English benchmarks Android Control, GUI-Odyssey, and AITZ~\cite{sun2024genesis, wu2024atlas, xu2024aguvis, lu2024gui, zhang2024aitz}. These benchmarks utilize two standard metrics: Type Match (TM), which verifies if the predicted action type matches the ground truth, and Exact Match (EM), which additionally requires all parameters to be correct. Table~\ref{tab:performance_android} showed that, for Android Control, our 3B model demonstrated significantly superior performance compared to a range of 7B models. The sole exception was a slightly lower score on the TM-Low metric, which we attribute to the inherent limitations of a 3B model in advanced instruction-following.
For GUI-Odyssey, our model achieved robust performance with only 3B parameters; all results are second only to the Odyssey-7B and Aguvis-7B models that underwent targeted training.
On AITZ, Mobile-R1 (3B) further outperforms every 7B baseline, reaching $77.05$ TM and $60.50$ EM, indicating that our three-stage recipe generalizes well beyond the Chinese apps seen during training.

Overall, these results underscore our model's robust and competitive performance on English-language tasks.


\begin{figure*}[t]
    \centering
  \includegraphics[width=1\linewidth]{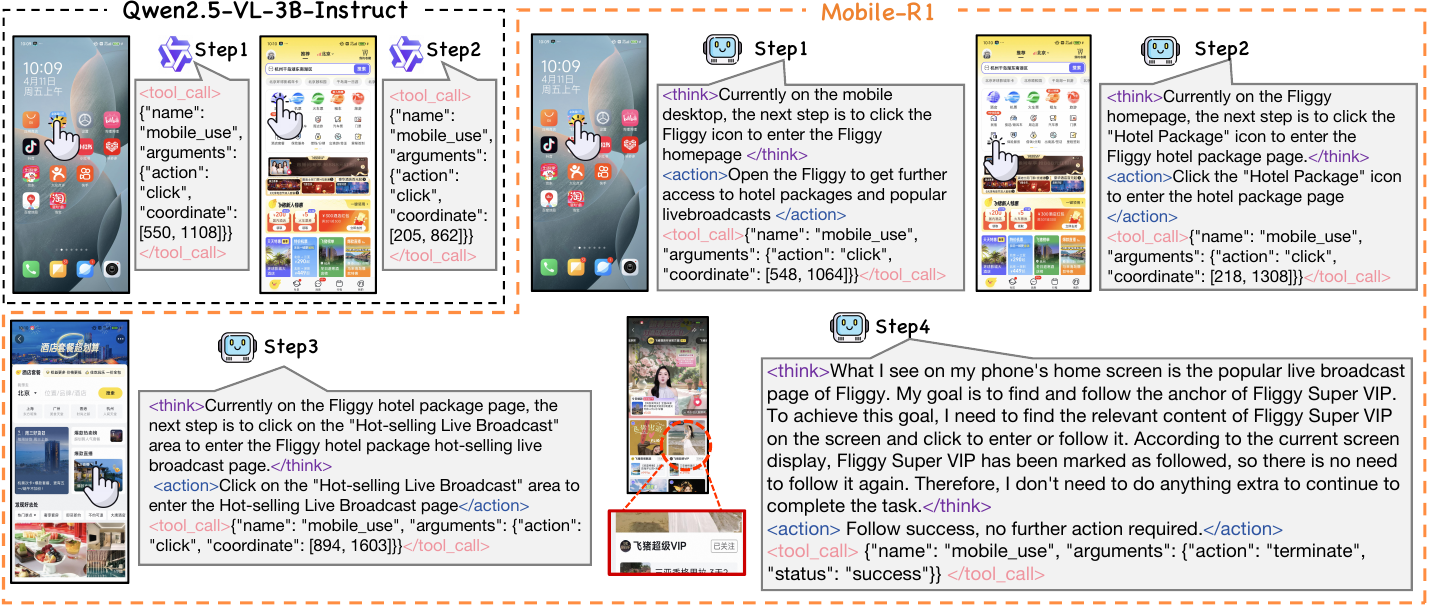}
    \vspace{-0.3cm}
    \caption{Comparison of reasoning trajectories between Mobile-R1 and Qwen2.5-VL-3B-Instruct on the task ``\textit{Open Fliggy, enter the hotel package, enter the popular live broadcast, find Fliggy Super VIP, and follow the anchor}''. In this case, Qwen2.5-VL-3B-Instruct failed at the second step, while Mobile-R1 completed the whole task accurately.}
    \label{fig:vis}
    \vspace{-0.2cm}
\end{figure*}

\subsection{Robustness Analysis}

\begin{figure}[t]
    \centering
  \includegraphics[width=0.95\linewidth]{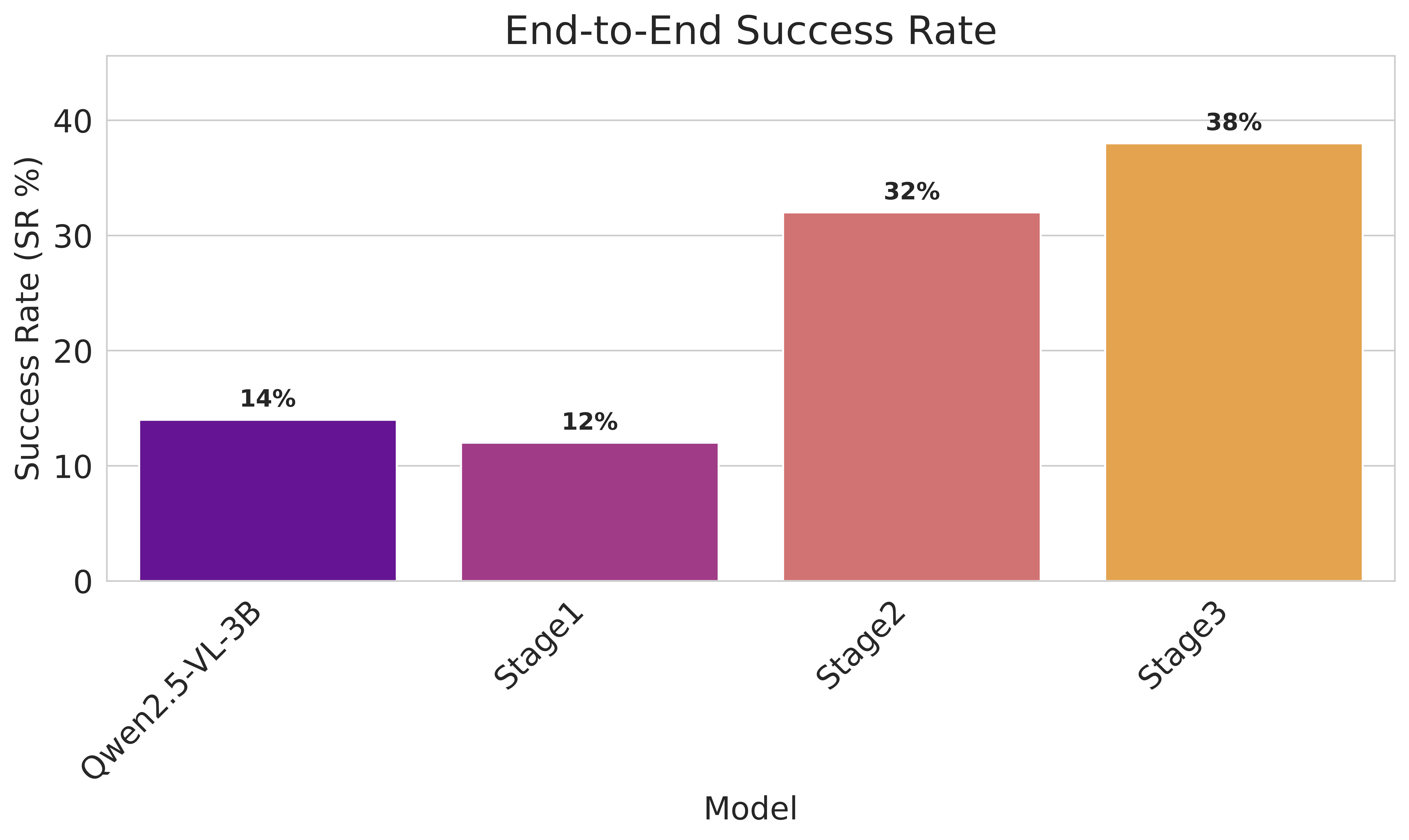}
    \caption{Task success on physical devices.}
    \label{fig:e2e}
    \vspace{-0.5cm}
\end{figure}

To assess the generalization capabilities of Mobile-R1, we validated our method's impact on end-to-end task performance by evaluating the success rate across 50 random tasks on physical devices. As detailed in Fiure~\ref{fig:e2e}, the model exhibited a consistent improvement trajectory. A minor initial dip in performance was observed during Stage 1, attributable to the action-level pre-training. Nevertheless, our method culminated in a significant 24 percentage point improvement in the final task success rate.

\subsection{Qualitative Visualization}

To effectively illustrate the performance of our Mobile-R1, we randomly selected several cases from the test set for qualitative analysis. 
As shown in Figure~\ref{fig:vis}, we have the following observations: 1) Qwen2.5-VL-3B-Instruct failed at the second step, while Mobile-R1 completed the entire task accurately.
2) At the second step, Qwen2.5-VL-3B-Instruct clicked the wrong icon labeled ``Hotel'' when the correct choice was ``Hotel Package.'' Mobile-R1 was able to click the correct icon, displaying precise icon identification.
3) At the final step, Mobile-R1 accurately recognized that the tab had already been marked as ``Followed,'' which eliminated the need for further actions, thus concluding the task with superior contextual awareness.
\section{Conclusion}
In this paper, we introduced a systematic training recipe and bridge the community’s need for high-quality Chinese GUI training and evaluation data.
Through our three-stage training process—including format finetuning, action-level GRPO training, and task-level GRPO training in realitic dynamic environments—Mobile-R1 significantly advances the capabilities of VLM-based mobile agents, demonstrating superior exploration and error correction abilities, and overcoming the limitations of prior methods that rely solely on single action prediction. Experimental results demonstrate that our Mobile-R1 surpasses all baselines in all metrics.

\section{Limitations} 
From the perspective of training strategy, while Mobile-R1 demonstrates promising capabilities in robust GUI interactionthis, our approach is essentially a systematic training recipe that employs both action-level and task-level rewards to guide the RL process, enabling the agent to improve progressively. It provides the interaction logic between GUI and environment for end-to-end training; however, we introduce no algorithmic innovation in RL itself. This will be a future direction: devising novel RL methods and refining reward design to achieve more efficient end-to-end interaction with mobile environments.
Regarding dataset, our final dataset, after rigorous quality filtering and difficulty balancing, represents only a small fraction of the overall data source.
Therefore, expanding the dataset size and incorporating more languages will be the focus of our future work.

Stage~3 of our Mobile-R1 uses a sliding window of size $W$ over historical screenshots (Eq.~\ref{eq:trajectory_def}; value in Appendix~\ref{appendix: training settings}) due to VRAM limits. The full \emph{textual} history is retained, which suffices for the self-correction behaviors we observe (Fig.~\ref{fig:intro}b, Fig.~\ref{fig:appvis}), but long-range \emph{visual} recall beyond $W$ steps is intrinsically restricted. Scaling $W$ via visual-token compression or external memory is left for future work.

\section{Ethical Considerations}
\textbf{Data Privacy:} Our dataset was collected from publicly available applications. We have strictly anonymized all sensitive Personally Identifiable Information (PII), such as phone numbers, addresses, and account details, during both the collection and annotation phases to protect user privacy.
\bibliography{custom}
\appendix

\newpage
\clearpage
\appendix\clearpage

\section{Overview of Appendix}
We have over 5 pages of this appendix, comprising the following subsections for the convenience of readers:

\begin{itemize}[leftmargin=*]
    \item \textbf{Appendix B}: More details of all prompts.
    \item \textbf{Appendix C}: More details of the dataset.
    \item \textbf{Appendix D}: More details of action space.
    \item \textbf{Appendix E}: More details of training settings.
\end{itemize}

\noindent \textbf{More experimental analysis}
\begin{itemize}[leftmargin=*]
    \item \textbf{Appendix F}: Robustness analysis of Mobile-R1.
    \item \textbf{Appendix G}: Analysis of Task-level Training.
\end{itemize}

\noindent \textbf{More visualization and cases}
\begin{itemize}[leftmargin=*]
    \item \textbf{Appendix H}: Visualization of performance comparison among models.
\end{itemize}

We hope that our efforts will serve as a source of inspiration for more readers!

\section{Prompts}
\label{appendix:prompts}
\subsection{A.1  Prompts of Dataset Generation}
\begin{figure}[!h]
    \centering
  \includegraphics[width=1.0\linewidth]{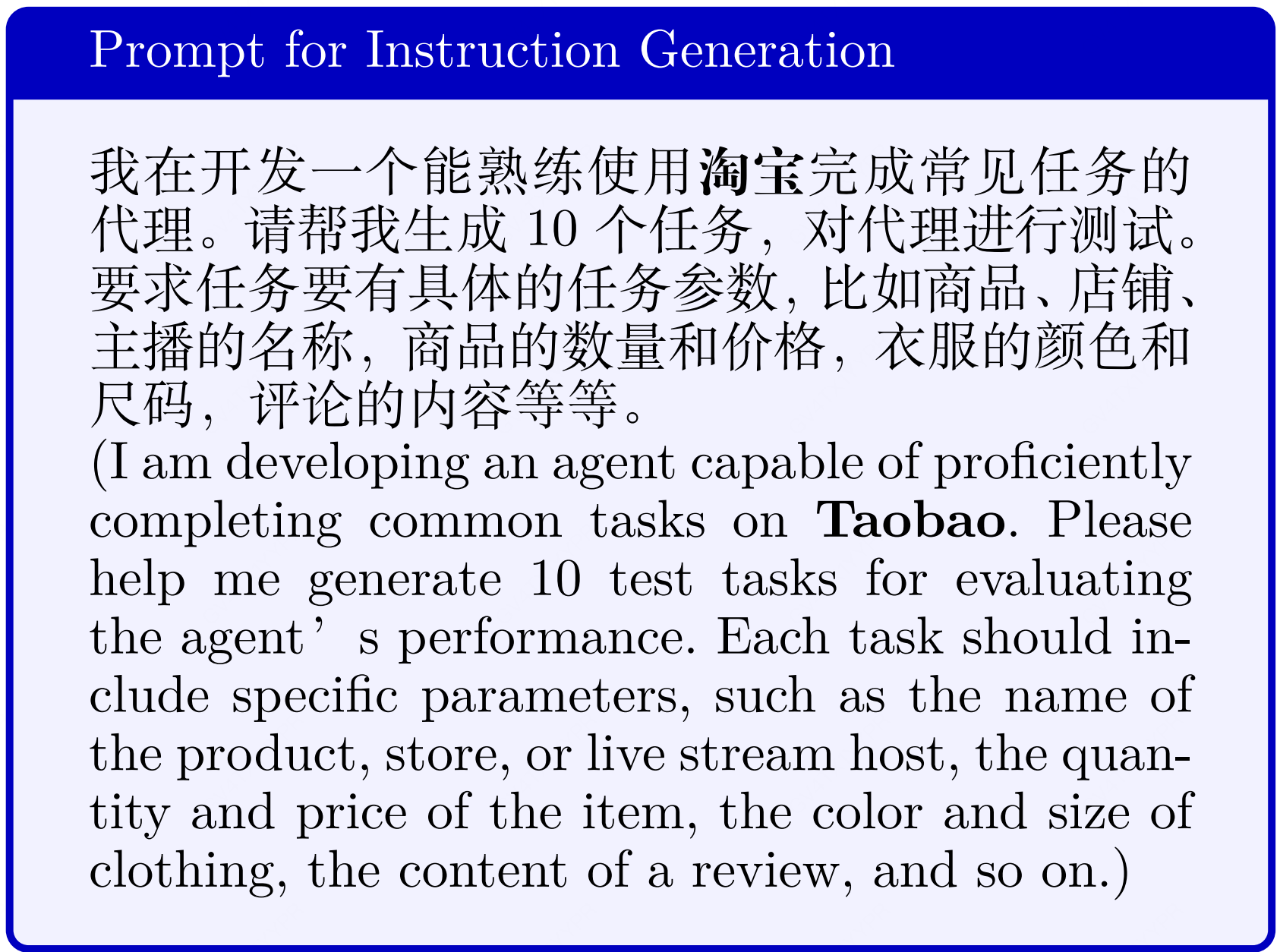}
    \caption{Prompt for Instruction Generation.}
    \label{fig:prompt_instruction}
\end{figure}

The prompt used to generate instructions for execution by mobile agents, derived from Claude 3.5, is shown in Figure~\ref{fig:prompt_instruction}. 
The \textbf{bond} indicates the name of an app , which can be replaced by any app listed in Table~\ref{tab:app_list}.

\subsection{Prompts of Mobile Agent}
The prompt used to guide the Qwen2.5-VL to execute the instructions for trajectory collection. The \textbf{bond} indicates the instruction, which can be replaced by any instruction of our dataset.

\begin{tcolorbox}[colback=yellow!5!white, colframe=yellow!50!black, title=Prompt for Agent Exploration]


You are a mobile GUI agent. You are given a task and your action history, with screenshots. You need to perform the next action to complete the task.\\

    You are provided with function signatures within \texttt{<tools></tools>} XML tags:

    \texttt{<tools>}

\begin{center}
\small 
\begin{verbatim}
{
  "name": "mobile_use",
  "arguments": {
    "type": "function",
    "function": {
      "name_for_human": "mobile_use",
      "name": "mobile_use",
      "description": "Use a touchscreen 
      to interact with a mobile device."
    }
  }
}
\end{verbatim}
\end{center}

    \texttt{<tools>}\\

    For each function call, return a json object with function name and arguments within \texttt{<tool\_call></tool\_call>} XML tags:
    
    \texttt{<tool\_call>}
\begin{center}
\small 
\begin{verbatim}
{
    "name": <function-name>, 
    "arguments": <args-json-object>
}
\end{verbatim}
\end{center}
    \texttt{</tool\_call>}\\

Analyze the task and historical actions, and predict the next step.

Output your reasoning process within the \texttt{<think></think>} tag.

Output the action to be performed in this step within the \texttt{<action></action>} tag.

Output the final answer within the \texttt{<tool\_call></tool\_call>} tag.\\

User Task: \textbf{Open Taobao and find the store I am following.}

\end{tcolorbox}

\subsection{Prompts of Judge Model}
\label{appendix:judge_prmopt}

\begin{figure}[!h]
    \centering
  \includegraphics[width=1.0\linewidth]{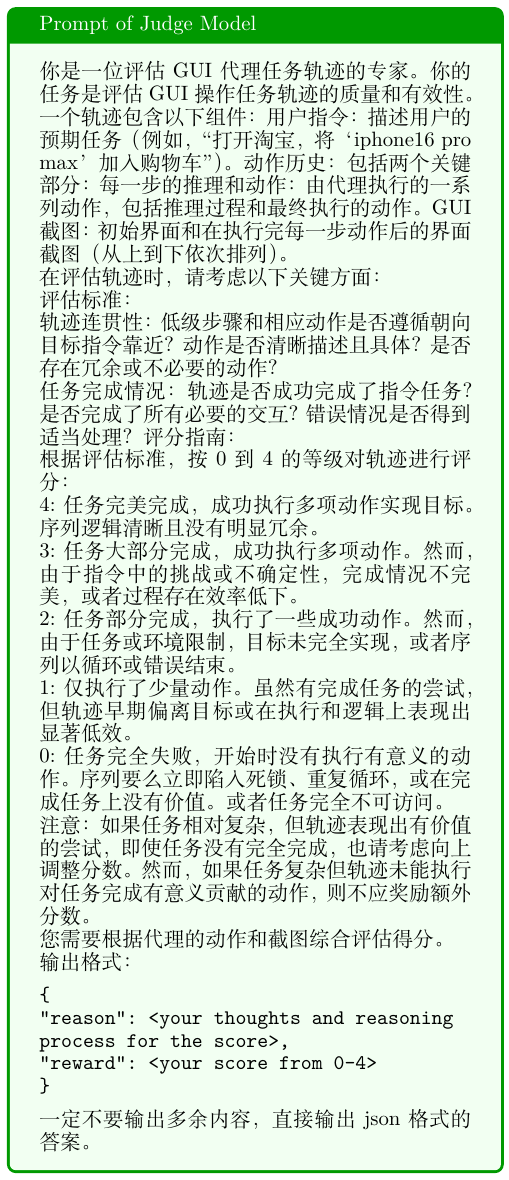}
    \caption{Prompt of Judge Model.}
    \label{fig:prompt_reward}
\end{figure}

As our experiments were conducted on Chinese applications, we accordingly prompted the scoring model in Chinese. The obtained scores were subsequently divided by 4 to normalize them into the $[0,1]$ range. The prompt is shown in Figure~\ref{fig:prompt_reward}.

\subsection{Prompts of Training System}
\label{appendix:train_prmopt}
This prompt was designed to ensure consistency with the Qwen2.5-VL's GUI prompt. The \textbf{bond} indicates the instruction, which can be replaced by any instruction of our dataset.

\begin{tcolorbox}[colback=purple!5!white, colframe=purple!60!black, title=Prompts of Training System]

You are a mobile GUI agent. You are given a task and your action history, with screenshots. You need to perform the next action to complete the task.\\

    You are provided with function signatures within \texttt{<tools></tools>} XML tags:

    \texttt{<tools>}

\begin{center}
\small 
\begin{verbatim}
{
  "name": "mobile_use",
  "arguments": {
    "type": "function",
    "function": {
      "name_for_human": "mobile_use",
      "name": "mobile_use",
      "description": "Use a touchscreen 
      to interact with a mobile device."
    }
  }
}
\end{verbatim}
\end{center}

    \texttt{<tools>}\\

    For each function call, return a json object with function name and arguments within \texttt{<tool\_call></tool\_call>} XML tags:
    
    \texttt{<tool\_call>}
\begin{center}
\small 
\begin{verbatim}
{
    "name": <function-name>, 
    "arguments": <args-json-object>
}
\end{verbatim}
\end{center}
    \texttt{</tool\_call>}\\

Analyze the task and historical actions, and predict the next step.

Output your reasoning process within the \texttt{<think></think>} tag.

Output the action to be performed in this step within the \texttt{<action></action>} tag.

Output the final answer within the \texttt{<tool\_call></tool\_call>} tag.\\

User Task: \textbf{Open Taobao, search for ``iPhone 16 Pro Max", and add it to the shopping cart.}

\end{tcolorbox}

\begin{table*}[ht]
\centering
\caption{Definition of Atomic Action Space}
\label{tab:action_space}
\renewcommand{\arraystretch}{1.2} 
\resizebox{0.99\textwidth}{!}{
\begin{tabular}{l l}
\toprule
\textbf{Action} & \textbf{Definition}          \\
\midrule
$key()$ & Perform a key event on the mobile device, supporting adb's `keyevent` syntax.    \\
$click(x, y)$  & Click the point on the screen with coordinate $(x, y)$.      \\
$swipe(x_{1}$, $y_{1}$, $x_{2}$, $y_{2})$ & Swipe from the starting point with coordinate $(x_{1}$, $y_{1})$ to the end point with coordinates2 $(x_{2}$, $y_{2})$.      \\
$long\_press(x, y, time)$  &  Press the point on the screen with coordinate $(x, y)$ for specified seconds.     \\
$type(text)$  &  Input the specified text into the activated input box.     \\
$system\_button(button)$  &  Press the system button, e.g., $Back, Home, Menu, Enter$.    \\
$terminate(status)$  &  Terminate the current task and report its completion status, e.g., $success, failure$. \\
$wait(time)$  &  Wait specified seconds for the change to happen. \\
\bottomrule
\end{tabular}}
\end{table*}

\section{More Details of the Dataset}
\label{appendix:dataset}

\subsection{App Statistics from the Dataset}
Our dataset includes 28 Chinese mobile applications, covering a wide range of daily-use scenarios. The detailed results are shown in Table~\ref{tab:app_list}.

\begin{table}[ht]
\centering
\caption{Statistics from Dataset}
\label{tab:app_list}
\renewcommand{\arraystretch}{1.2} 
\begin{tabular}{r l r r}
\toprule
\textbf{\#} & \textbf{App} & \textbf{Data} & \textbf{Trajectory} \\
\midrule
1  & Alipay          & 5    & 1   \\
2  & Amap            & 446  & 93  \\
3  & App Store       & 16   & 4   \\
4  & Baidu           & 439  & 106  \\
5  & Baidu Maps      & 1844 & 308  \\
6  & Bilibili        & 2457 & 376 \\
7  & Browser         & 213  & 39   \\
8  & Calculator      & 443  & 56  \\
9  & Calendar        & 390  & 69  \\
10 & Dianping        & 4    & 1   \\
11 & Douyin          & 447  & 106  \\
12 & Eleme           & 538  & 129  \\
13 & Fliggy          & 2934 & 593  \\
14 & Idle Fish       & 29   & 5   \\
15 & JD              & 1092 & 211  \\
16 & Kuaishou        & 2938 & 619 \\
17 & Luckin Coffee   & 144  & 18   \\
18 & Meituan         & 520  & 93  \\
19 & Mobile System   & 38   & 7   \\
20 & Notes           & 110  & 19  \\
21 & Pinduoduo       & 707  & 115  \\
22 & Quark           & 390  & 67  \\
23 & Taobao          & 4326 & 821 \\
24 & Tencent Maps    & 912  & 177  \\
25 & WeChat          & 33   & 6   \\
26 & Weather         & 396  & 91  \\
27 & Xiaohongshu     & 2706 & 504 \\
28 & Zhuanzhuan      & 4    & 1   \\
\midrule
   & \textbf{Total}  & 24,521 & 4,635 \\
\bottomrule
\end{tabular}
\end{table}

\subsection{App Statistics from Open-Source Data}

To facilitate the training and utilization of our dataset, focusing on Chinese mobile applications, we have open-sourced a sample of 1007 trajectories covering 28 different applications. Notably, for reasons related to data review, we have selected a portion of the data for open access. In the coming months, we will gradually organize and release additional data. The detailed results are shown in Table~\ref{tab:app_list_open}.

\begin{table}[ht]
\centering
\caption{Statistics from Open-Source Data}
\label{tab:app_list_open}
\renewcommand{\arraystretch}{1.2} 
\begin{tabular}{r l r r}
\toprule
\textbf{\#} & \textbf{App} & \textbf{Data} & \textbf{Trajectory} \\
\midrule
1  & Alipay          & 3    & 1  \\
2  & Amap            & 227  & 50 \\
3  & App Store       & 11   & 4  \\
4  & Baidu           & 160  & 52 \\
5  & Baidu Maps      & 191  & 50 \\
6  & Bilibili        & 148  & 55 \\
7  & Browser         & 174  & 39  \\
8  & Calculator      & 195  & 50 \\
9  & Calendar        & 162  & 50 \\
10 & Dianping        & 4    & 1  \\
11 & Douyin          & 152  & 51 \\
12 & Eleme           & 193  & 50 \\
13 & Fliggy          & 191  & 50  \\
14 & Idle Fish       & 27   & 5  \\
15 & JD              & 200  & 50  \\
16 & Kuaishou        & 136  & 51 \\
17 & Luckin Coffee   & 97   & 18  \\
18 & Meituan         & 252  & 50 \\
19 & Mobile System   & 7    & 3  \\
20 & Notes           & 64   & 19 \\
21 & Pinduoduo       & 180  & 51  \\
22 & Quark           & 181  & 50 \\
23 & Taobao          & 221  & 50 \\
24 & Tencent Maps    & 231  & 50 \\
25 & WeChat          & 33   & 6  \\
26 & Weather         & 159  & 50 \\
27 & Xiaohongshu     & 296  & 50 \\
28 & Zhuanzhuan      & 4    & 1  \\
\midrule
   & \textbf{Total}  & 3,924 & 1,007 \\
\bottomrule
\end{tabular}
\end{table}





\section{Atomic Action Space}

Table~\ref{tab:action_space} presents all atomic operations considered in our framework. There are eight actions: key, click, swipe, long press, type, system button, terminate, and wait.

\begin{figure*}[t]
    \centering
  \includegraphics[width=1\linewidth]{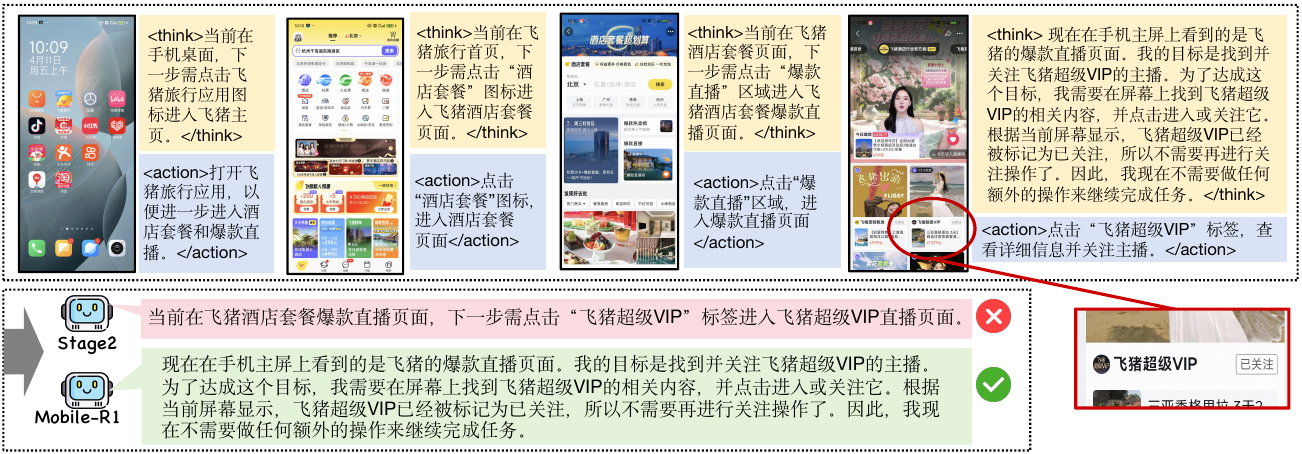}
    \caption{Comparison of the thinking processes of the Stage2 model and Stage3 model of Mobile-R1 under the same task ``\textit{Open Fliggy, enter the hotel package, enter the popular live broadcast, find Fliggy Super VIP, and follow the anchor.}"}
    \label{fig:appvis}
\end{figure*}

\section{Training Settings}
\label{appendix: training settings}
Our experiments utilize Qwen2.5-VL-3B as the base model, with GRPO implementation (including for trajectory-level interaction training) adapted from the open-r1 framework.

\textbf{For action-level training:}
\begin{itemize}
    \item Supervised Fine-Tuning (SFT): LoRA was applied for SFT, training for 2 epochs with a learning rate of $1 \times 10^{-4}$. The mini-batch size per device was set to 1, and a gradient accumulation number of 8 was used.
    \item GRPO Training: This phase involved 2 epochs of training with a learning rate of $1 \times 10^{-7}$. A mini-batch size of 1 per device and a gradient accumulation number of 2 were configured. To encourage exploration, the number of rollouts was set to 8, and the temperature was set to 1.
\end{itemize}

\textbf{For trajectory-level training:}
\begin{itemize}
    \item We utilized two parallel virtual machine instances running locally to conduct experiments, enabling real-time interaction with the simulated environment.
    \item The number of rollouts was set to 4, and the maximum exploration steps per interaction was limited to 14. This allowed the model to explore and potentially backtrack historical operations upon error detection.
    \item Training was conducted for 2 epochs with a learning rate of $1 \times 10^{-6}$. A temperature of 1 was used to encourage exploration.
    \item A gradient accumulation number of 4 was applied, and due to the continuous real-time interaction with the virtual environment, a mini-batch size of 1 per device was maintained.
    \item Our Android emulator was configured to a Medium Phone device profile. To balance computational efficiency, each screenshot (window size $W$ of Eq \ref{eq:trajectory_def}) fed to the model was downsampled by a factor of 2 in both dimensions.
\end{itemize}

\section{Robustness analysis of Mobile-R1}
\label{appendix: robutness analysis}
\begin{figure}[t]
    \centering
  \includegraphics[width=0.95\linewidth]{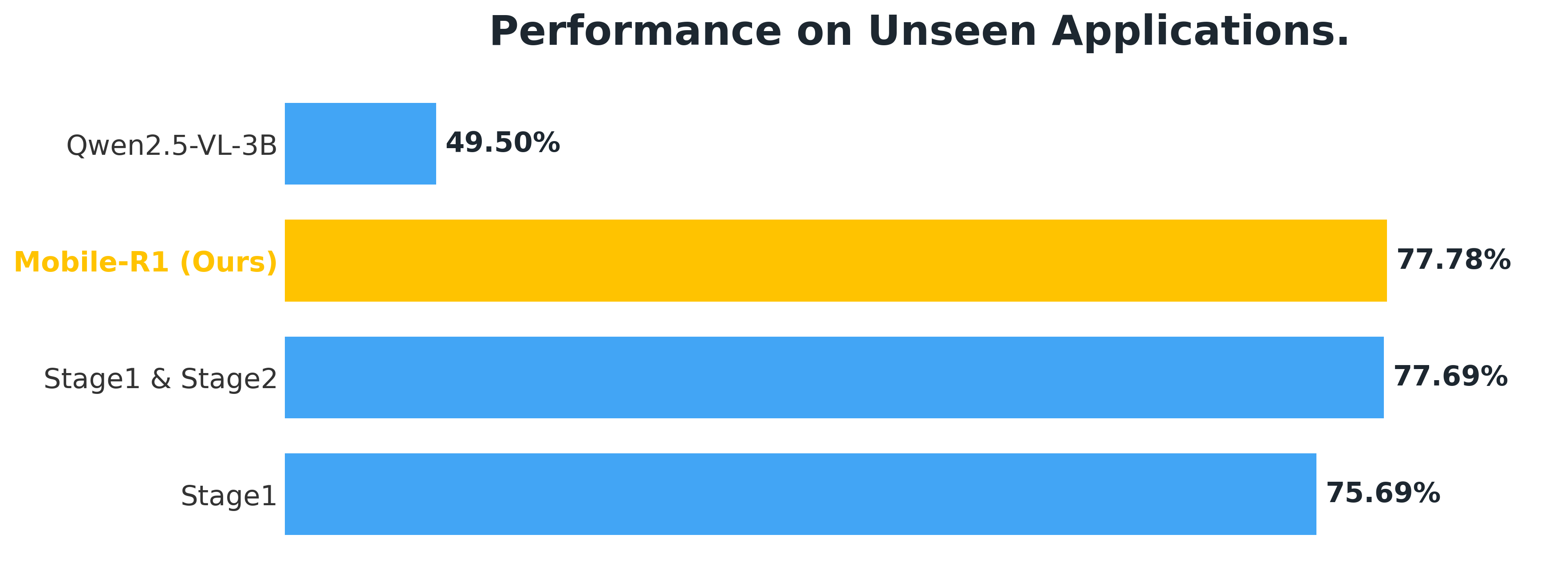}
    \caption{Robustness Analysis on Unseen Applications.}
    \label{fig:unseen}
    \vspace{-0.5cm}
\end{figure}

We performed a robustness analysis on 225 trajectories from unseen apps.
The experimental results are shown in Figure~\ref{fig:unseen}. The following observations can be made: 
1) All of the Models exhibits a noticeable decline in accuracy, indicating challenges in its robustness and generalization.
2) Mobile-R1 demonstrates the best performance in scores, emphasizing its enhanced generalization capabilities. 
This improvement is largely attributable to the crucial role of Stage 3 training, which significantly bolsters both robustness and adaptability.

\begin{figure}[t]
    \centering
  \includegraphics[width=0.95\linewidth]{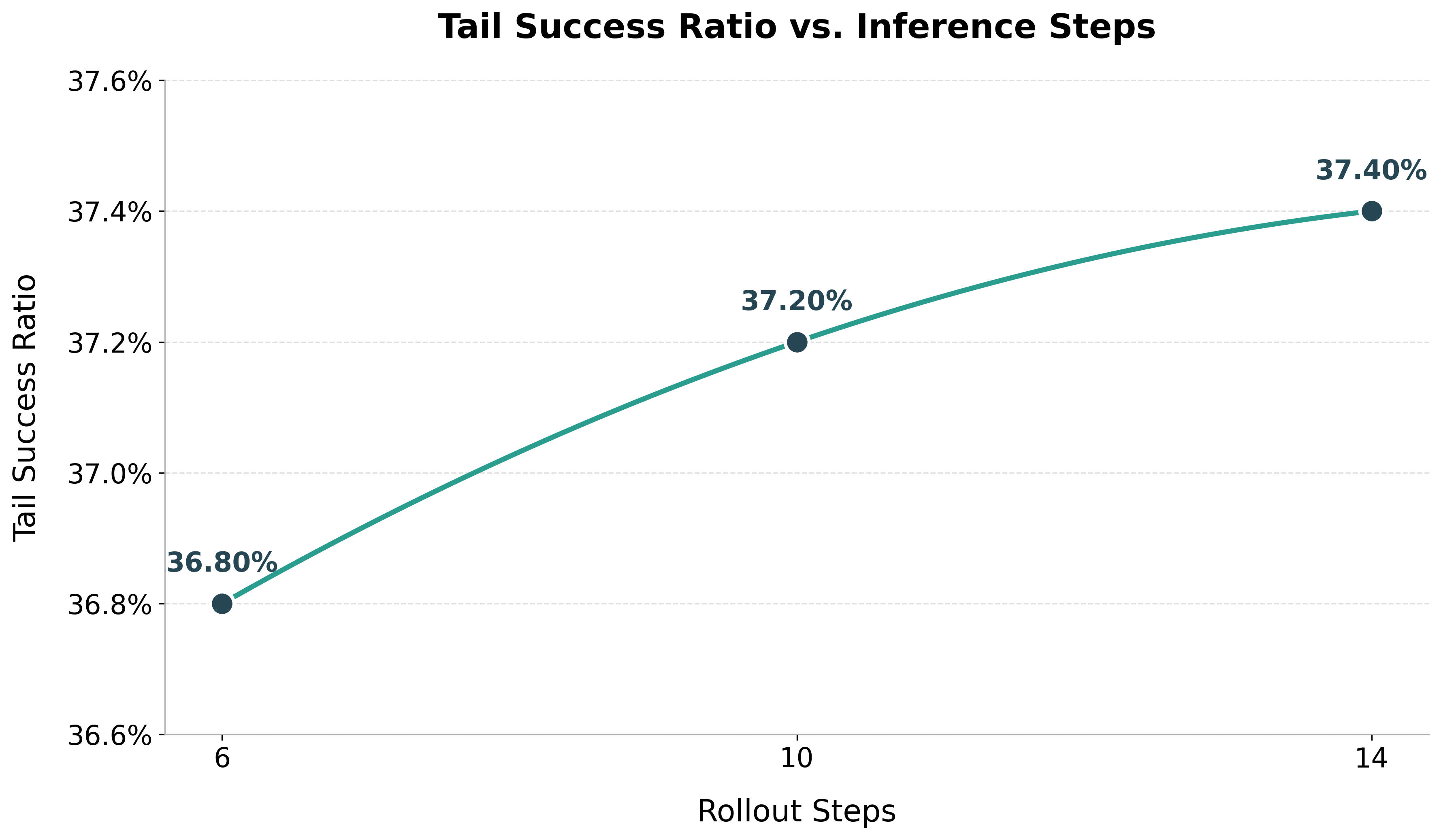}
    \caption{Tail success rate on different inference steps.}
    \label{fig:steps}
    \vspace{-0.5cm}
\end{figure}

We further investigated the impact of varying the number of training steps in Stage 3. Specifically, we tracked the model's tail success ratio as a function of the training steps. As illustrated in Figuree~\ref{fig:steps}, the tail success ratio exhibits a consistent upward trend with the increase in steps. This finding suggests that a more prolonged phase of end-to-end exploration is beneficial for achieving final task success.

\begin{figure}[t]
    \centering
  \includegraphics[width=0.95\linewidth]{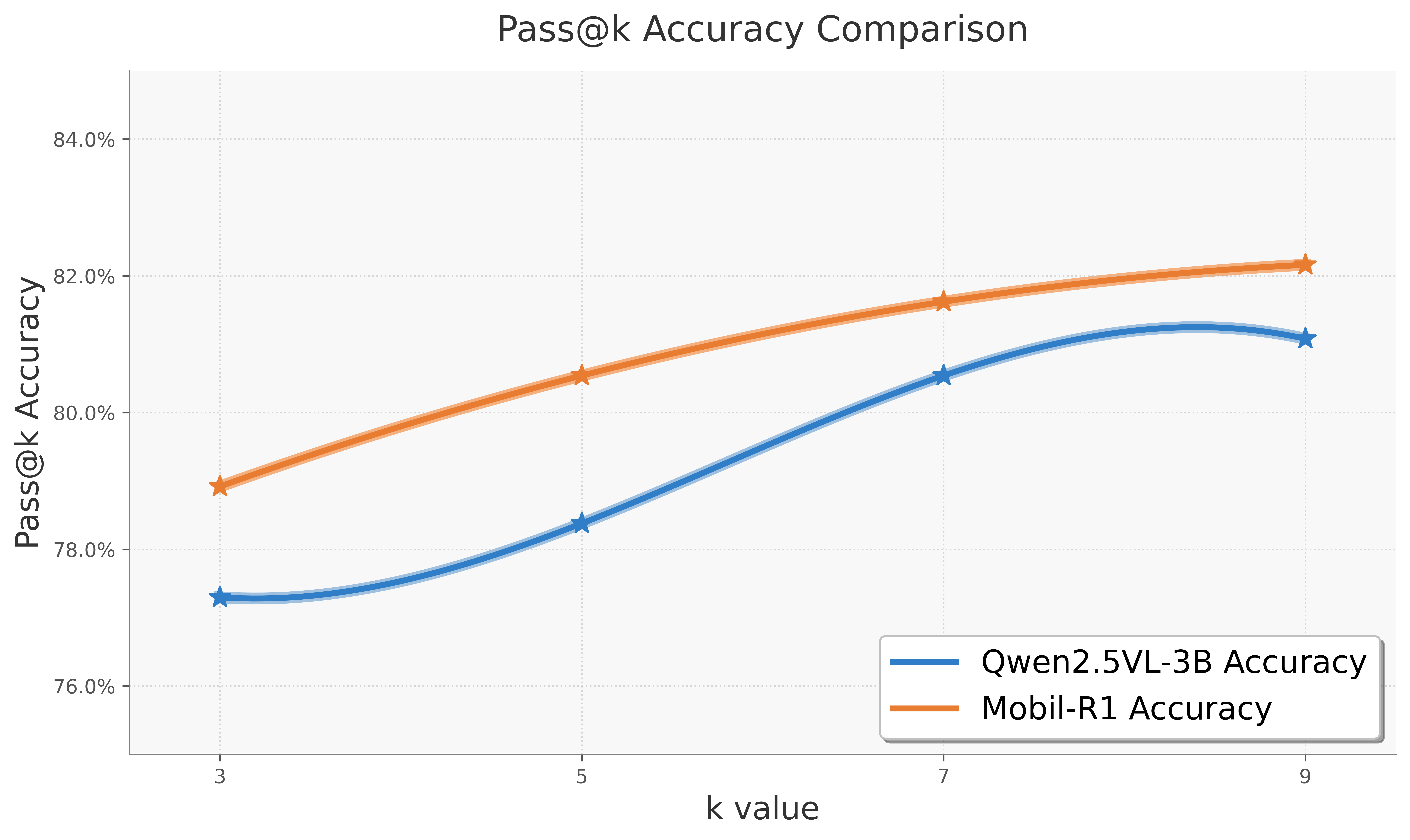}
    \caption{Pass K performance of Mobile-R1.}
    \label{fig:passk}
    \vspace{-0.5cm}
\end{figure}

To comprehensively unveil the upper bounds and maximum potential of our model's performance, we studied how model accuracy changes with pass@k — a metric that defines the probability that at least one of the k attempts successfully solves the given problem.
We tested our models on 50 randomly sampled complete trajectories (185 actions) from our test set.
During inference, we used a temperature of 0.7. Figure \ref{fig:passk} shows both models' accuracy increases significantly as k grows, highlighting their ability to solve problems with multiple attempts. 
Notably, our Mobil-R1 model exhibited superior performance compared to the baseline model, achieving an absolute accuracy improvement of up to 7 percentage points.

\section{Analysis of Task-level Training}
\label{appendix: task-level training}
\begin{figure}[t]
    \centering
  \includegraphics[width=0.95\linewidth]{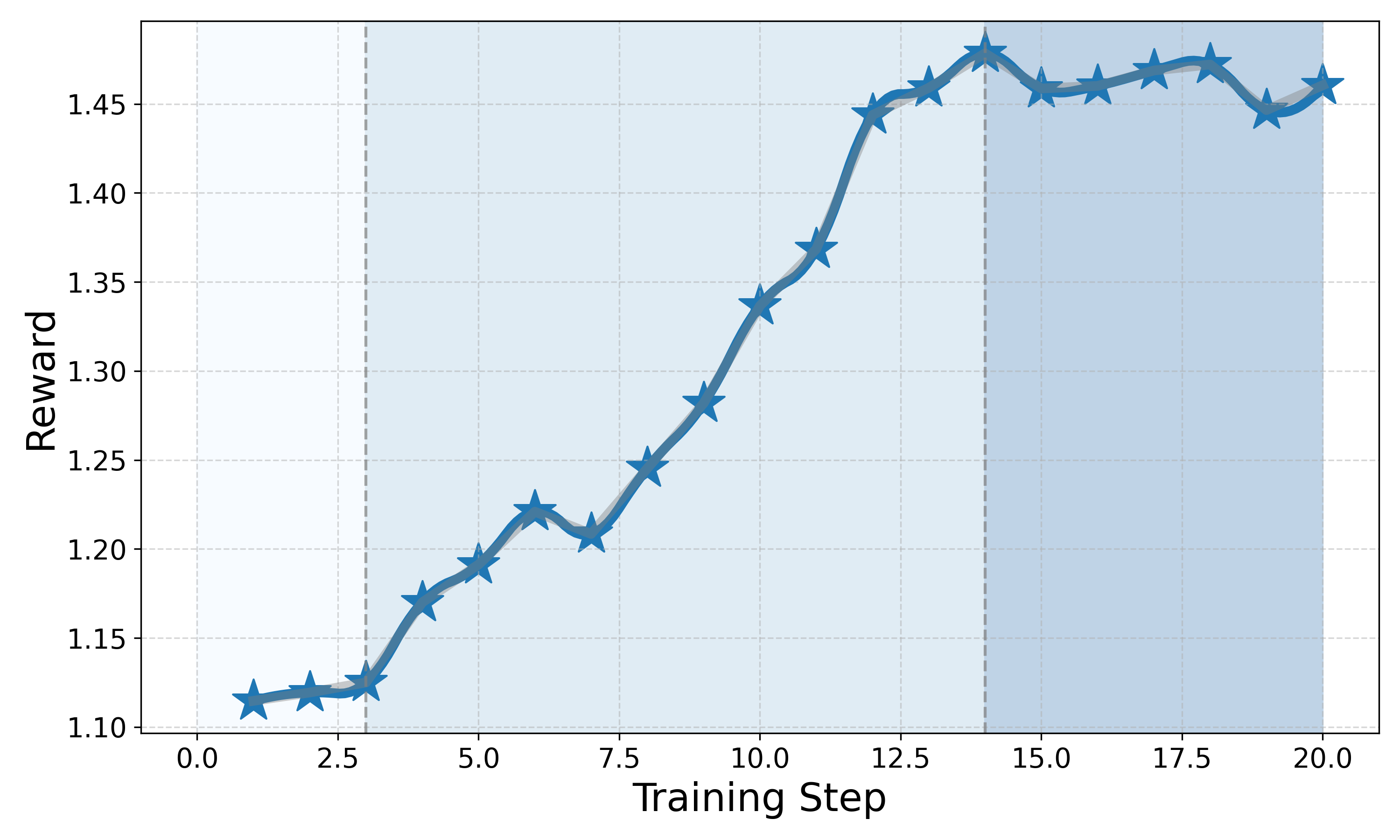}
    \caption{Reward score during Stage3 training.}
    \label{fig:reward_stage3}
\end{figure}
The reward score of Stage3 shown in Figure~\ref{fig:reward_stage3}, reveals slow reward growth in the first three training steps, likely due to instability caused by aggressive exploration or policy changes.
However, between steps 4 and 14, the reward growth accelerates, suggesting effective learning. Subsequently, from step 15 onward, the growth stabilizes, indicating that the policy is gradually converging.

Overall, Figure~\ref{fig:reward_stage3} shows that despite the initial
instability induced by aggressive exploration, the policy successfully
stabilizes and converges within two epochs. This two-epoch schedule
is therefore a deliberate trade-off that balances exploration depth
against computational cost, and the observed plateau after step~15
justifies stopping there rather than training for additional epochs.

\section{More Qualitative Results}
Figure~\ref{fig:appvis} demonstrates the comparison of the thinking processes between the Stage2 model and Stage3 model of Mobile-R1 under the same task.
At the final step of the task, Mobile-R1 accurately recognized that the tab was already marked as ``Followed'', eliminating the need for further actions and thus concluding the task with superior contextual awareness. In contrast, the Stage 2 model still attempted to click on the ``Feizhu Super VIP'' tab, demonstrating a lack of awareness of the task's completion status.

Notably, this Eureka moment does not require revisiting earlier screenshots: the ``Followed'' cue that triggers termination is fully contained in the Step~4 screenshot, while the memory of prior actions is carried via the textual trajectory. This is a concrete instance of the two dependency sources identified in \S\ref{sec:stage3}, and explains why self-correction remains effective under a bounded visual window~$W$.

\begin{figure*}[t]
    \centering
  \includegraphics[width=0.85\linewidth]{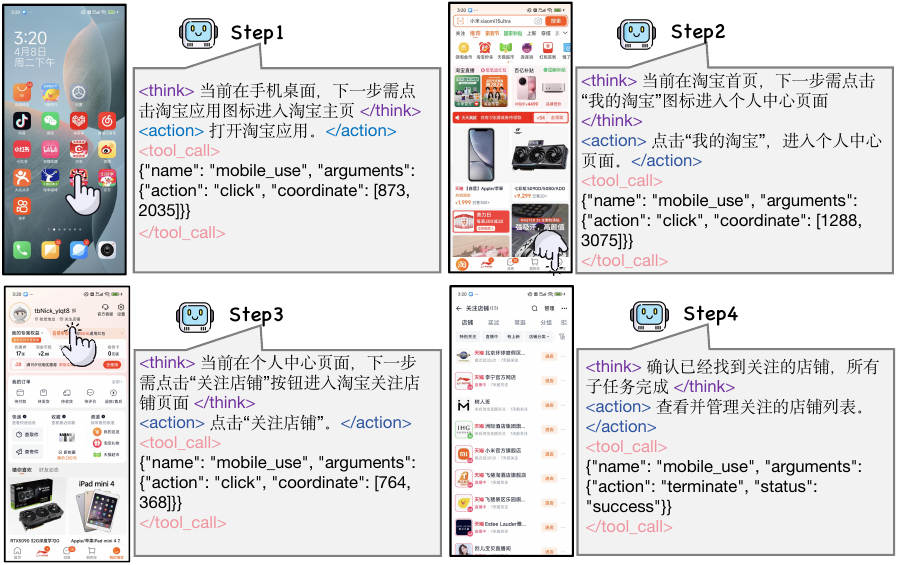}
    \caption{Qualitative result of Mobile-R1 under the task ``\textit{Open Taobao and find the followed shops.}"}
    \label{fig:case_1}
\end{figure*}

Figure~\ref{fig:case_1} demonstrates the successful trajectory of the Mobile-R1 agent in completing the task ``\textit{Open Taobao and find the followed shops.}"

\begin{figure*}[t]
    \centering
  \includegraphics[width=1\linewidth]{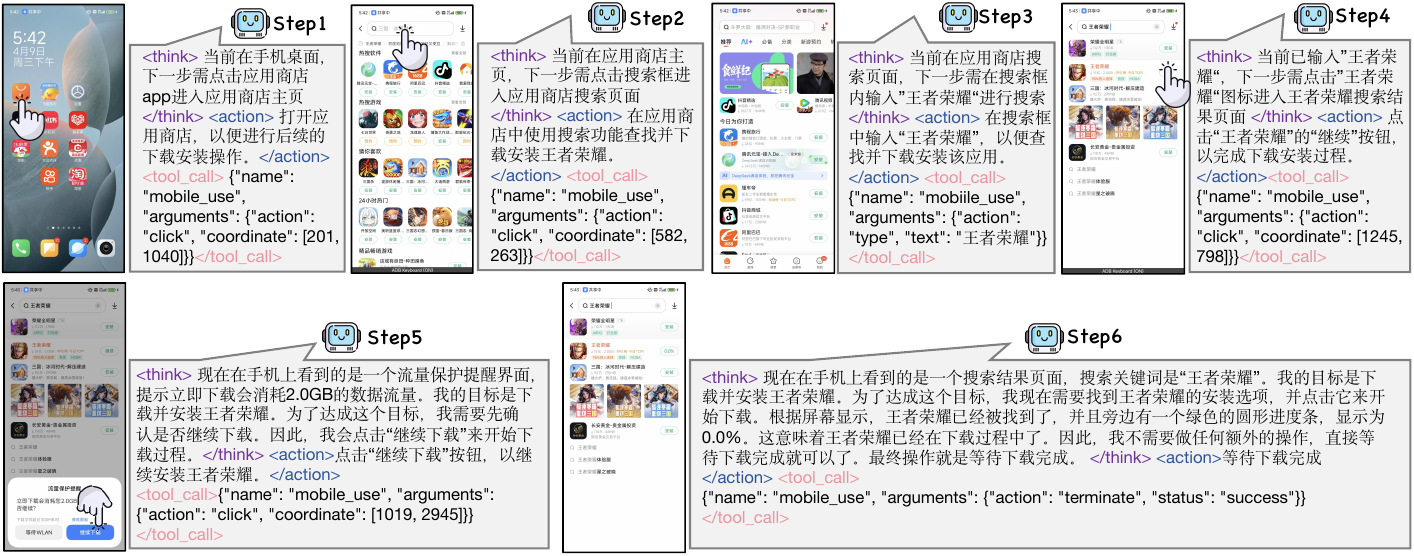}
    \caption{Qualitative result of Mobile-R1 under the task ``\textit{Download and install Honor of Kings.}"}
    \label{fig:case_2}
\end{figure*}

Figure~\ref{fig:case_2} demonstrates the successful trajectory of the Mobile-R1 agent in completing the task ``\textit{Download and install Honor of Kings.}"

\section{Human Validation of the GPT-4o Judge}
\label{app:human_eval}
 
To verify that GPT-4o is a trustworthy trajectory-level judge for Stage~3, we performed a blind human evaluation.
 
\paragraph{Setup.}
We randomly sampled $30$ trajectories previously scored by GPT-4o, stratified across the five rubric levels ($0$--$4$; see Appendix~\ref{appendix:judge_prmopt}). Three human annotators --- blind to the GPT-4o scores --- independently re-scored each trajectory using the \emph{same} rubric, viewing only the screenshots, thinking traces, and actions. The annotator majority label was taken as the human score.
 
\paragraph{Results.}
Exact-level agreement was $21/30$ ($70.0\%$), and the remaining $9$ cases each differed by exactly one level, so $|\Delta|\!\le\!1$ agreement reached $30/30$ ($100\%$) with a mean absolute deviation of only $0.30$ levels. The overall ranking was well preserved: Spearman $\rho = 0.92$ and Kendall $\tau_b = 0.86$, with a quadratic-weighted Cohen's $\kappa = 0.92$ on the ordinal five-level rubric.
 
\paragraph{Discussion.}
These results support the reasonableness of using GPT-4o as the trajectory-level judge. Importantly, Stage~3 uses the \emph{group-normalized} advantage in Eq.~(\ref{eq:grpo_objective}): absolute scoring errors are partially cancelled by the group mean, so what primarily drives the policy gradient is the \emph{relative} ordering within a rollout group, which the high rank correlation directly validates. Small boundary-level discrepancies therefore have limited impact on the learning signal. Three design choices further stabilize the judge: (i) the five-level rubric reduces scoring variance; (ii) the prompt (Fig.~\ref{fig:prompt_reward}) fixes two evaluation dimensions --- \emph{Trajectory Coherence} and \emph{Task Completion} --- to mitigate drift; and (iii) scores are re-normalized to $[0,1]$ before being used as rewards.

\section{All Resources}
We are committed to releasing all our resources, including the dataset, model weights, and framework implementation, which are slated for public release, to foster reproducibility. The project homepage is available at: https://mobile-r1.github.io/Mobile-R1/. To adhere to the double-blind review process, the resource links on the page are currently placeholders. They will be made fully public upon acceptance of the paper.

\end{document}